\documentclass{article}

\usepackage{PRIMEarxiv}

\usepackage[utf8]{inputenc} 
\usepackage[T1]{fontenc}    
\usepackage{hyperref}       
\usepackage{url}            
\usepackage{booktabs}       
\usepackage{amssymb}
\usepackage{amsmath}
\usepackage{multirow}
\usepackage{amsfonts}       
\usepackage{nicefrac}       
\usepackage{microtype}      
\usepackage{lipsum}
\usepackage{fancyhdr}       
\usepackage{graphicx}       
\graphicspath{{media/}}     

\title{Integrating knowledge from case reports: a medical ontology based multimodal
information system with structured summary
}

\author{
  Shuyu Guo \\
  College of Computer Science and Technology \\
  Jilin University \\
  Changchun\\
   \And
  Lan Huang \\
  College of Computer Science and Technology \\
  Jilin University \\
  Changchun \\
   \And
  Yichen Liu \\
  College of Software \\
  Jilin University \\
  Changchun \\
   \And
  Hanbin Ma \\
  Department of Laboratory Medicine \\
  Sichuan Provincial People's Hospital, School of Medicine, \\University of Electronic Science and Technology of China \\
  Chengdu \\
   \And
  Tian Bai* \\
  College of Computer Science and Technology \\
  Jilin University \\
  Changchun \\
}

\begin{document}
\maketitle

\begin{abstract}
Published medical case reports serve as a crucial medical information carrier, documenting discoveries in rare diseases, diagnostic methods, and innovative treatments. Despite the wealth of clinical knowledge in millions of case reports in the public medicine literature database (PubMed), accessing relevant information efficiently is hindered by the limitations of traditional keyword-based retrieval tools on unstructured and diverse case reports. To address the above issues, we introduce a comprehensive multimodal information system for case reports integrating structured clinical summaries of patients including medical images and biomedical named entities from 52949 open-access case reports published from 2000 to 2021. The multimodal essential information is organized in a well-structured medical ontology. Also, a powerful interface for searching and browsing case reports is designed to assist junior clinicians in retrieving cases effectively and improving the identification and diagnosis of rare diseases.

\end{abstract}

\keywords{medical case report \and multimodal information integration \and literature retrieval \and named entity recognition \and data mining}

\section{Introduction}
In evidence-based medicine, structured and well-formulated materials bring clinicians more valuable evidence and boost efficiency of retrieving appropriate resources for medical treatment \cite{b1,b2}. As a prominent type of medical published literature, case reports detailedly record the first discovery of patients with rare diseases, symptoms, special clinical prognoses, or novel treatments to achieve the share of clinical experiences and knowledge in the practice of medicine with clinicians worldwide. Case reports integrate optimal research evidence with clinical expertise and case values, which are significant sources of pedagogical reference. Facing abnormal conditions in diagnosis, it is time-consuming for clinicians, especially those with less clinical experience, to read case reports completely to assist their treatments. To focus on this limitation, constructing a comprehensive database with high-quality and structured multimodal summaries from case reports is of high necessity.

\begin{figure}[!t]
\centerline{\includegraphics[width=0.6\columnwidth]{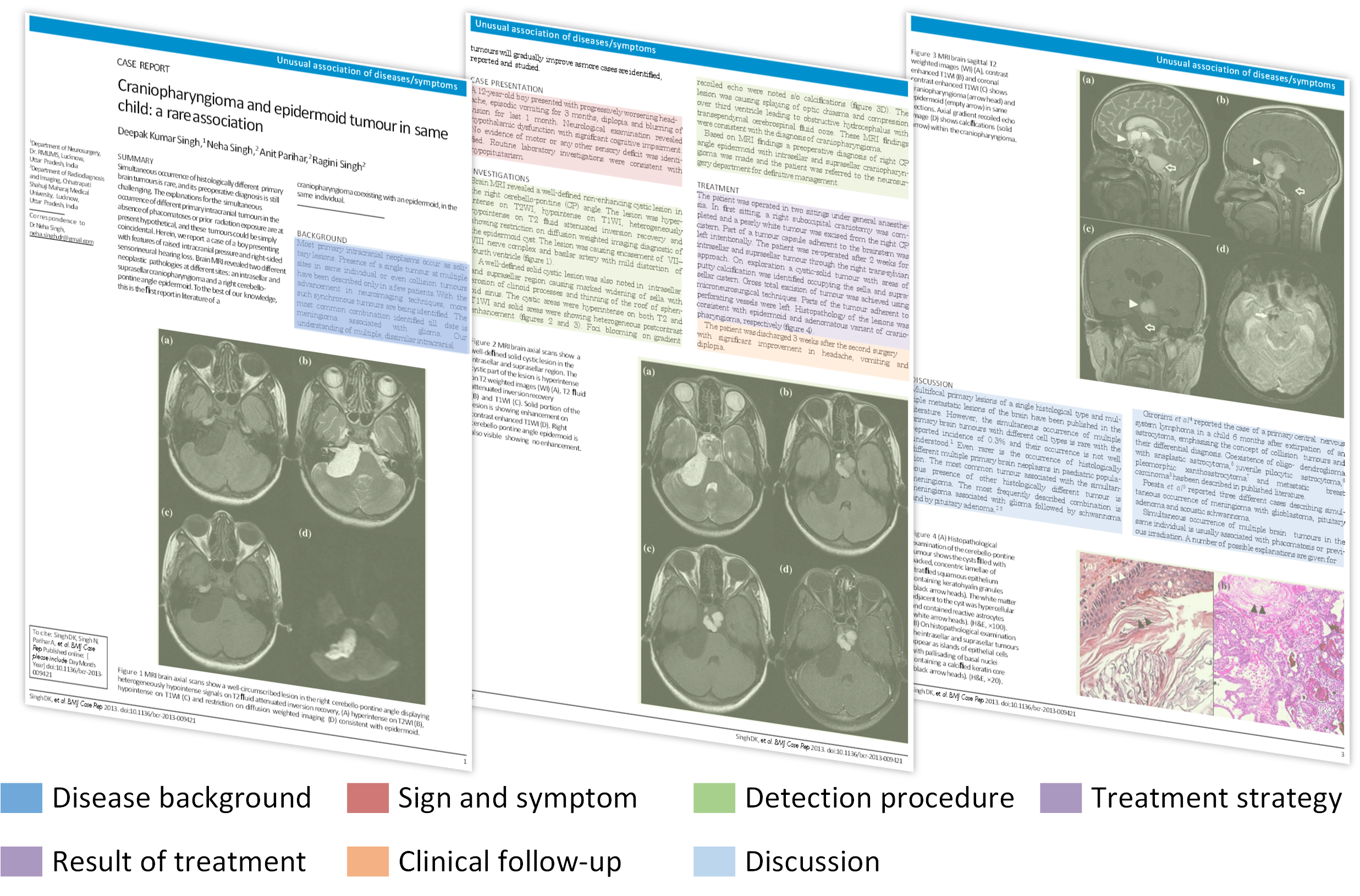}}
\caption{An example of main sections contained in the case report.}
\label{fig1}
\end{figure}

In Egyptian period, clinicians wrote down medical knowledge acquired from practical experience on antiquity papyrus as clinical case notes \cite{b3}. By the end of the 18th century, case notes had evolved into standardized case description essays and been organized into sections mainly including general patient information, history, examination details, treatment, and subsequent course of condition \cite{b4,b5}. In 1893, the public open-access medicine literature database, PubMed Central (PMC), began to collect case essays as official medical literature named case reports. Since then, editable full texts of medical literature have been available, which contributes greatly to the automatic extraction on case reports. Millions of case reports following formatted contents \cite{b6,b7} have been collected sharing medical knowledge across the worldwide healthcare community. The standard template of a case report usually contains 5 main sections \cite{b8} (Figure \ref{fig1}), which could be summarized as: (i) sign and symptom of patient; (ii) detection procedure on patient; (iii) treatment strategy; (iv) result of treatment; (v) clinical follow-up.

Recently, studies on clinical knowledge have mainly focused on electronic medical records (EMR), where case description is written in a fixed pattern to improve clinical and healthcare efficiency from an operational standpoint. Researchers have developed methods to extract features from EMR and built a clinical support system \cite{b9, b10} for more personalized clinical care. Accordingly, a series of computational methods \cite{b11, b12, b13} are proposed to address issues on clinical description extraction and patient risk prediction based on deep learning models. Edward et al. proposed a graph convolutional transformer that could jointly learn the hidden structure of electronic records to solve the problem that the structure information was not available \cite{b14}. Matthieu et al. extracted implicit knowledge from EMR data of patients by developing a reinforcement learning agent \cite{b15}. Most researchers have focused on extracting clinical information from EMR texts such as patients’ histories, current symptoms, conditions, and biomarkers, which could assist in diagnosis and lead to the development of more targeted clinical interventions \cite{b16,b17,b18}. A free accessible EMR database, integrating a decade of detailed information about individual patient care, was established by Johnson et al. \cite{b19}, which has promoted the development of some methods for analysis and prediction on EMR \cite{b20}, for instance, de-identification to patient EMR privacy protection \cite{b21}, prediction of clinical events \cite{b22}, interactive semantic search on clinical records \cite{b23}. Considerable achievements have been made in extracting valuable information from EMR to assist clinicians, however, lower accessibility results from ethical and privacy authorization requirements from hospitals, weaker representatives, and less reference value owing to massive collections of all patient cases without any screening process. On the contrary, case reports are published as medical literature with the advantages of large quantities, high availability, and strong educational guidance, which have not been sufficiently utilized yet.

Along with the ever-increasing amounts of open-access biomedical papers, literature retrieval is no longer restricted to titles and abstracts but also includes full text. Volanakis et al. provided a literature search service to improve literature retrieval accuracy by using cited statements carrying information in abstracts and full texts \cite{b24}. Alodadi et al. linked the clinical entities extracted from electronic records with the biomedical literature to optimize doctors’ diagnosis \cite{b25}. Demner et al. developed a multimodal biomedical information retrieval system named Open-i that enables the retrieval of abstracts and figures from free-access literature and biomedical image collections from libraries and hospitals \cite{b26}. Research on literature retrieval has achieved great performance, however, few efforts focus on case reports. Luo et al. constructed a machine learning-based model \cite{b27,b28} to automatically extract main findings from abstracts and full texts on the basis of a manually annotated corpus \cite{b29} of main finding sentences. Even though a brief description of final conclusion or diagnosis is included in the main finding sentences of case reports, a substantial amount of clinical information (i.e. symptoms of patients, laboratory tests, and medical imaging figures) scatters over other sections of the full text. Some clinicians prefer obtaining patient conditions through medical images due to their direct expression. Open-i also provides retrieval on case report figures by matching textual keywords with captions and abstracts, however, clinicians could be uncertain about the specific disease names facing patients with unusual phenotypes and need to review case reports by matching similar clinical symptoms. The accuracy of search results depends on whether the query terms are close to the topics. Most existing literature retrieval methods require exact keywords related to the title or topics, which are unworkable in case of unknown diseases. Building a comprehensive medical case report database allows clinicians to efficiently retrieve knowledge based on structured clinical summaries to support their diagnosis, especially under the situation of indistinction for some rare diseases.

To improve the identification and diagnosis of rare diseases, open knowledge sharing, and data structuring are crucial to assist clinicians. The focus of this study is to extract multimodal summaries (i.e. medical imaging figures, clinical and biological entities) from medical case reports and apply disease ontology, symptom ontology, and body system ontology for reconstructing case summaries to build multimodal database named CRFinder. Additionally, a user-friendly interface is designed for clinicians to retrieve and analyze case reports by using medical ontology filters and browsing medical imaging figures, which could provide junior clinicians hints when they are unaware of a specific name or treatment of a rare disease. The contributions of this study can be summarized as follows:

(1) To summarize key information, a comprehensive structured multimodal database focusing on medical case reports, CRFinder, is first introduced, which comprises structured case summaries including medical entities and medical images of different modalities. 

(2) Medical ontologies of diseases, symptoms, and body systems are applied for reconstructing multimodal summaries. We design a novel retrieval pattern on case reports by browsing extracted medical figures, which provides clinicians with medical ontology-based hints and improves the identification of rare diseases and unexpected associations among diseases and symptoms. 

(3) CRFinder also provides a user-friendly web-based retrieval platform to assist clinicians with efficient visualization and analysis of different modalities of essential information extracted from case reports. Two retrieval functions are developed including medical figures browsing and keyword searching cater to clinicians’ retrieval preference for rare diseases.

\begin{figure*}[!t]
\centering{\includegraphics[width=\textwidth]{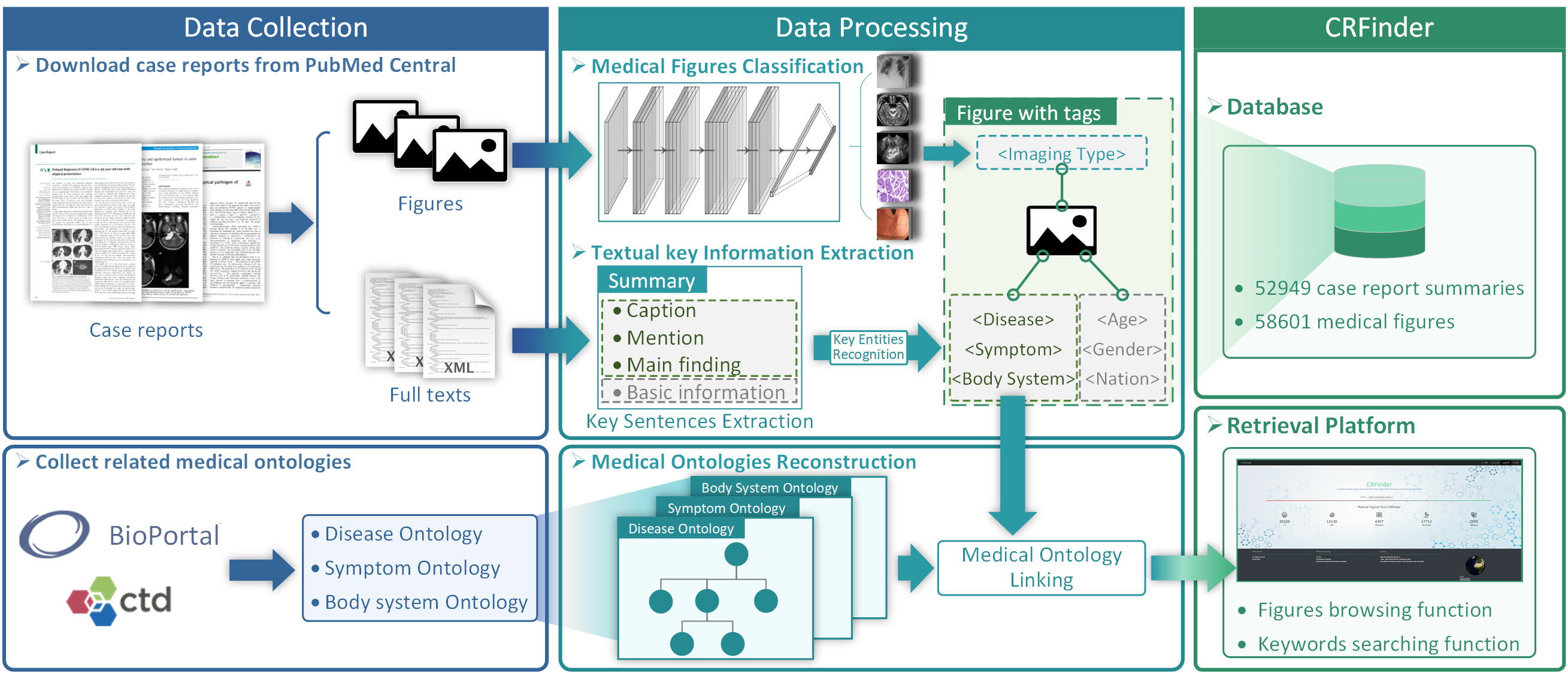}}
\caption{An overview of CRFinder construction. \textbf{Data Collection:} Download Case Reports from PMC; Construct medical ontologies of disease, symptom and body system based on CTD and BioPortal. \textbf{Data Processing:} Classifying figures based on clinical imaging type; Extract textual summaries, and recognize key entities as figure tags; Link figures with tags based on medical ontologies. \textbf{CRFinder:} Build structured multimodal database of case reports; Design a retrieval platform with two retrieval functions.}
\label{fig2}
\end{figure*}

\section{Material and methods}
\subsection{Data collection}
Millions of human case reports are collected in different medical publishing companies and literature databases, which have become a valuable resource for clinical knowledge. To obtain high quality and well-formatted source data, case reports published from 2000 to 2021 written in English are selected and downloaded from open access subset of PMC. Owing to some articles having copyright protection, not all articles in PMC are available for text mining or other reuse. Finally, 52949 Case Report files are collected through FTP server from PMC\footnote{https://ftp.ncbi.nlm.nih.gov/pub/pmc/} with existing free licenses. Each downloaded file includes full text in XML format and figures in JPG format. Vocabulary from the Comparative Toxicogenomics Database (CTD\footnote{http://ctdbase.org/}, updated on 19 October 2020) is applied as disease ontology for constructing disease entities in CRFinder database. Symptom Ontology and Human Developmental Anatomy Ontology, obtained from comprehensive repository of biomedical ontologies (BioPortal)\footnote{https://bioportal.bioontology.org}, are also served for organizing symptom and body system entities. Figure \ref{fig2} provides a complete workflow of general protocol for CRFinder construction.

\subsection{Textual key information extraction}
For the database construction of CRFinder, the textual key information in case reports contains basic information and clinical information. Basic information describes patient background about age, gender, and nation. Clinical information describes the patient's clinical conditions of disease, symptoms and body system, laboratory results, imaging procedures, and final diagnosis. Based on the large amount of case reports we have reviewed, the textual key information usually appears in sentences such as caption sentences, figure-mentioned sentences, case description sentences, and main finding sentences, which are defined as key sentences of case reports in this study. For better operation on these sentences, Natural Language Toolkit (NLTK) package is applied to split full texts into multiple sentence tokens for filtering out the above key sentences as follows:

\textbf{Caption sentences:} Captions can be regarded as direct textual descriptions of figures, most of which include imaging type of figures, body system of imaging, and description of abnormal area. In the XML files downloaded from PMC, captions have been already annotated by the tags of ‘\textless caption\textgreater \textless /caption\textgreater’, which allows screening the caption sentences by the rule-based method.

\textbf{Mention sentences:} In case some of the captions are too simple to extract useful information, mentions are applied as supplements to enrich textual descriptions corresponding to the figures. Mentions usually express more detailed information about figures and appear in the main body of case reports explaining the meanings of figures. Mention sentences can be located by identifying keywords (e.g., \emph{fig} and \emph{figure}) in the full texts.

\textbf{Case description sentences:} According to the writing standard \cite{b31}, all published case reports should provide basic information about patients such as age, and gender, which usually appear in the first sentence of the case description paragraph. This pattern allows us to extract age, gender, infected location, or nation of the patient as basic information by recognizing sentences containing like “\emph{A 25-year-old male....}”.

\textbf{Main finding sentences:} Generally, case reports would indicate the novelty for their publishing, which could be a discovery of rare disease or syndrome, a novel and effective treatment, and so on. It is found that \cite{b27}, the sentences of main contribution descriptions prefer to begin with the templates like, “\emph{We report....}”, “\emph{Therefore,....}” or “\emph{In conclusion....}” appearing in the conclusion section or the end of the abstract. We regard these sentences as main finding sentences. However, using the above rule-based extracting method may collect some irrelevant sentences that deviate from the topic of the case report. To address this issue, Medical subject headings (MeSH terms), are adopted to further filter out main finding sentences. MeSH terms are special topic words assigned by subject matter experts to each article in PubMed and these terms could reflect the main topics of a medical article. We calculate the similarity scores between these rule-based extracted main finding sentences and case report topics. Specifically, we apply word co-occurrences of MeSH terms and main finding sentences to score each sentence and select the top three with the highest scores as final selections. If the main finding sentence contains more MeSH terms, it would be given a higher score. The similarity scores of main finding sentences are calculated as follows:

\begin{equation}
S_{MeSH}=\left \{ MT_{1},MT_{2},MT_{3},....MT_{n} \right \}
\label{eq1}\end{equation}

\begin{equation}
score=\begin{cases}
\frac{num_{co-occur}}{n-num_{co-occur}}   & \text{ if } num_{co-occur}<n \\
1  & \text{ if } num_{co-occur}=n
\end{cases}
\label{eq2}\end{equation}

\noindent where ${S_{MeSH}}$ is the MeSH term ${{MT}_i}$ set of one case report. ${n}$ is the number of MeSH terms assigned in one case report. ${{num}_{co-occur}}$ represents the number of MeSH terms occurring in main finding sentences. For each case report, main finding sentences with the top three scores would be collected in CRFinder database, which summarizes the rareness and main contributions of case reports. In addition, some metadata of case reports such as paper PubMed unique identifier (PMID), title, keywords, MeSH terms, abstract, and link of the original paper are also included in CRFinder database.

\subsection{Medical imaging type recognition}
According to our research, figure types in case reports are mainly as follows: (i) Medical imaging figures that represent disease under different clinical procedure imaging such as computed tomography (CT scanning), magnetic resonance imaging (MRI), ultrasonography, microscopy, endoscopy; (ii) Other scientific figures that indicate some data statistics, such as bar charts, scatter plots, and line charts. As a direct visualization of clinical knowledge, medical imaging figures in case reports allow clinicians to quickly understand the clinical conditions of patients.

We classify figures from case reports in two ways: (1) Select keywords in corresponding captions as figure labels; (2) Train a model for automatic label prediction. In the first way, based on the extracted caption sentences, we focus on keywords such as CT, MRI, ultrasound, microscopy, endoscopy, and any other synonyms to be figure labels. This keyword-based filter could annotate figures in an unsupervised and fast way. Secondly, using figures annotated in the first way as datasets to train deep learning models. We choose several baseline models for image classification including residual network (ResNet) \cite{b32}, dense connection network (DenseNet) \cite{b33}, inception network (Inception V3) \cite{b34}, and efficient network (EfficientNet) \cite{b35}. The model with the best performance would be utilized to classify the rest unlabeled medical imaging figures. In total, 7788 figures (including 1600 CT figures, 1540 MRI figures, 1303 endoscopy figures, 1600 microscopy figures, and 1745 ultrasound figures) are randomly selected as training datasets, and 1966 figures are selected as test datasets. All figures are resized to 300*300 dimensions to meet the input requirement of each model. The implementation details of each model are consistent. Adam optimizer and cross-entropy loss function are set in our experiments. Other parameter settings are as follows: batch size is 16, learning rate is 10e-3, and epoch is 30. All models are pre-trained on ImageNet. The accuracy of each model is shown in the Result section.

\begin{figure}[!ht]
\centering{\includegraphics[width=0.8\textwidth]{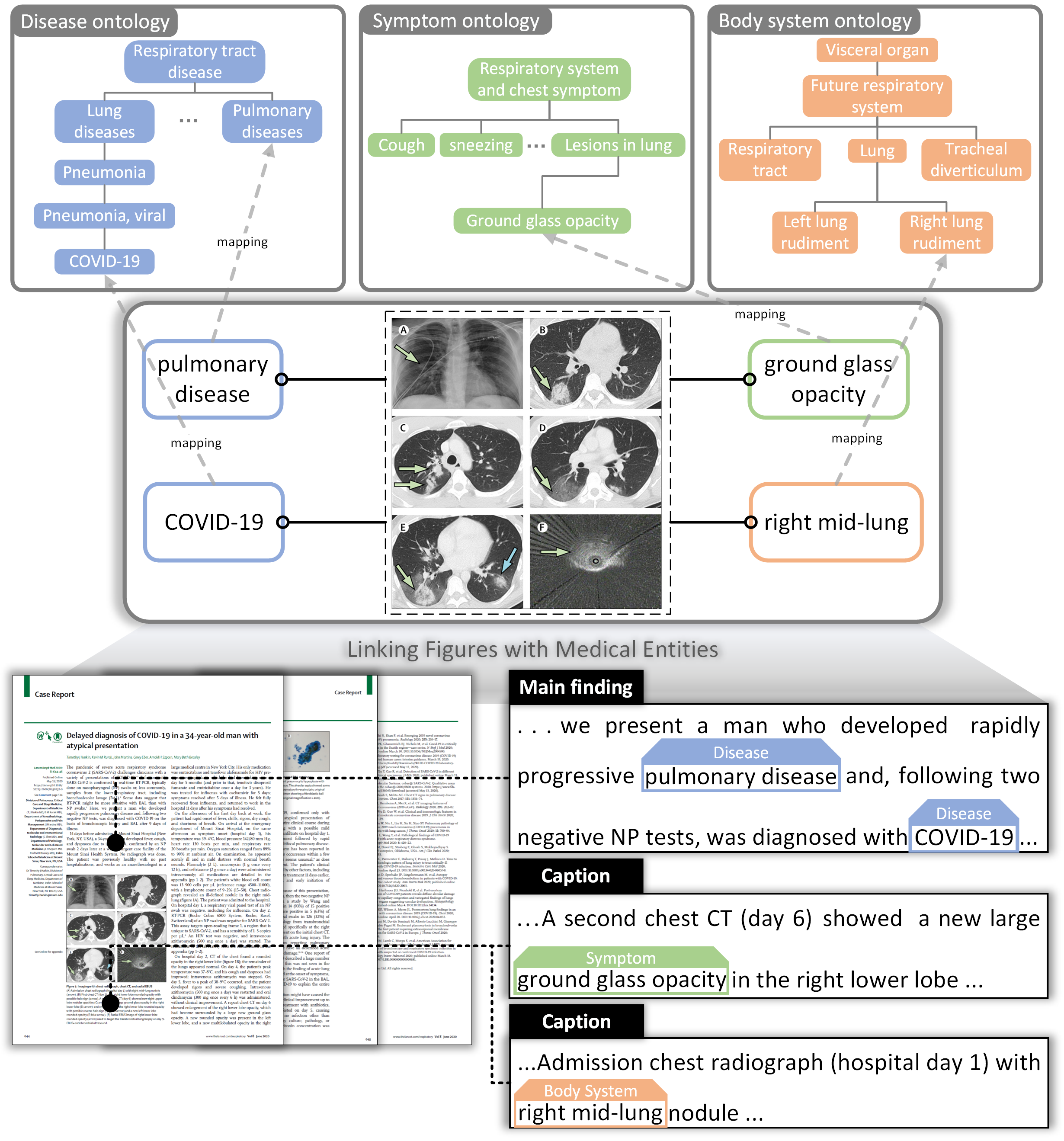}}
\caption{An example of medical entities extraction on key sentences and linking to medical ontologies of disease, symptom, and body system.}
\label{fig3}
\end{figure}

\subsection{Linking figures with entities based on medical ontologies}
In case reports, textual information provides clinical descriptions in detail and medical imaging figures illustrate patient lesions in a straightforward visual way. Figures and texts complement each other in describing patient clinical conditions. To bridge figures and medical entities and organize the extracted information, we construct CRFinder database in two ways: (i) Apply medical imaging figures as index anchors of each item in CRFinder database; (ii) Link figures with medical entities based on medical ontologies. Medical entities extracted from key sentences would be linked to figures as multiple tags for better organizing based on medical ontologies. Each item in CRFinder database includes one medical imaging figure, related entities (such as disease, symptom, body system, and biological entities if had), related sentences, and other figures in the same Case Report if has (displayed in thumbnails). Figure \ref{fig3} presents an example \cite{b36} of our linking pipeline.

We apply QuickUMLS\footnote{https://github.com/Georgetown-IR-Lab/QuickUMLS}, which is an unsupervised named entities recognition tool, for biomedical named entities extraction from case reports. Based on the unified semantic types \cite{b38} in Medical Language System (UMLS) and clinician-concerned clinical concept types, we mainly focus on the entity types of disease, symptom, and body system, which are shown in Table \ref{table1}.

\begin{table}[h!]
\centering
\caption{Selected biomedical entity types in UMLS.\label{table1}}%
{%
\begin{tabular}{cllll}
\cline{1-2}
Entity Type                  & \multicolumn{1}{c}{UMLS Semantic Type}     &  &  &  \\ \cline{1-2}
Disease                      & T047: Disease or Syndrome                  &  &  &  \\ \cline{1-2}
\multirow{4}{*}{Symptom}     & T019: Congenital Abnormality               &  &  &  \\
                                & T037: Injury or Poisoning                  &  &  &  \\
                                & T046: Pathologic Function                  &  &  &  \\
                                & T184: Sign or Symptom                      &  &  &  \\ \cline{1-2}
\multirow{6}{*}{Body System} & T022: Body System                          &  &  &  \\
                                & T023: Body Part, Organ, or Organ Component &  &  &  \\
                                & T024: Tissue                               &  &  &  \\
                                & T029: Body Location or Region              &  &  &  \\
                                & T030: Body Space or Junction               &  &  &  \\
                                & T031: Body Substance                       &  &  &  \\ \cline{1-2}
\end{tabular}%
}
\end{table}

The above three types of biomedical entities are extracted from key sentences and linked to disease ontology, symptom ontology, and body system ontology for a better-structured organization. To avoid the impact of naming conventions and synonyms during the linking process, we first employ Metathesaurus, a large biomedical thesaurus organized by concept or meaning, to normalize each extracted entity. Then, apply the IDs (CUI) to link biomedical entities with the nodes in medical ontologies. Furthermore, besides entity types in Table \ref{table1}, biological entities (such as gene, chemical, and protein) are also extracted as a supplement for CRFinder by using an automatic biomedical key concept annotation tool (Pubtator) \cite{b39}.

Based on the above textual medical entity and figure extraction and linking process, each case report is summarized into figures with multiple tags (biomedical entities) and organized by medical ontologies. Organizing CRFinder database in the form of medical ontologies can provide hierarchical prompts in cases where clinicians are unaware of rare diseases.

\section{Results}
\subsection{Statistics of CRFinder database}
\begin{figure*}[!t]
\centering{\includegraphics[width=0.9\textwidth]{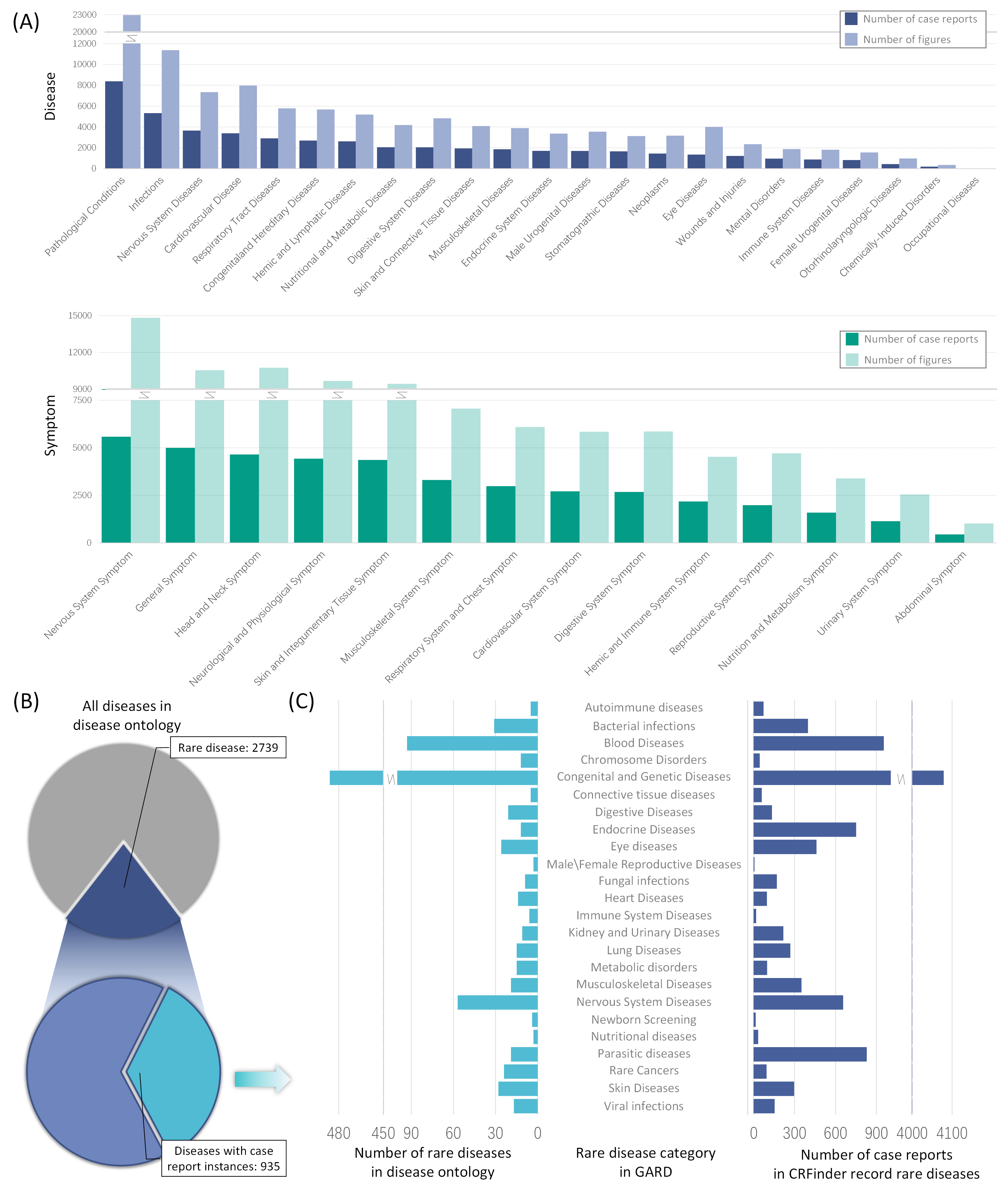}}
\caption{(A) Case reports and figure numbers under main root nodes of disease ontology and symptom ontology. (B) Distribution of rare diseases in disease ontology and contained in CRFinder database. (C) Case reports and disease numbers in rare disease category of GARD.}
\label{fig4}
\end{figure*}

The current version of CRFinder database contains 55212 different patient cases extracted from 52949 case reports in English published from 2000 to 2021. After classifying figures, 58601 of them are recognized as medical images including 30200 CT figures, 12136 MRI figures, 4357 ultrasound figures, 17712 microscopy figures, and 2595 endoscopy figures. Some of the figures may have multiple labels as they are compound images. Each figure is annotated by at least one kind of clinical entity. Currently, CRFinder database contains 5538 disease entities, 3101 symptom entities, and 4929 body system entities. Based on these entity annotations, figures are further linked to the medical ontologies of diseases, symptoms, and body systems. CRFinder database covers 55212 patients across different ages ranging from 1 year old to centenarian. More details are presented in Table \ref{table2} and Figure \ref{fig4}.

\begin{table*}[t]
\caption{Data statistics of 10 main body systems in CRFinder database. Numbers in Disease and Symptom columns respectively indicate the disease and symptom types in the corresponding body system. The last 5 columns indicate the numbers of 5 common clinical imaging types under each body system.\label{table2}}
\centering
{%
\begin{tabular}{cccccccc}
\hline
\multirow{2}{*}{Body System} & \multirow{2}{*}{Disease} & \multirow{2}{*}{Symptom} & \multicolumn{5}{c}{Numbers of different imaging figures} \\ \cline{4-8} 
                                &                          &                          & CT      & MRI   & Ultrasound  & Microscopy  & Endoscopy  \\ \hline
limb                         & 200                      & 51                       & 899     & 559   & 158         & 735         & -          \\
cardiovascular system        & 951                      & 30                       & 8218   & 3176  & 1117        & 4519        & 639       \\
integumental system          & 1084                     & 66                       & 1500    & 691   & 263         & 1713        & 140        \\
nervous system               & 3209                     & 209                      & 5144    & 3275  & 705         & 2830        & 305        \\
sensory organ                & 2027                     & -                        & 2431    & 1038  & 313         & 1430        & 163        \\
skeleton muscular            & 1927                     & 52                       & 2831    & 930  & 160         & 785        & -        \\
alimentary system            & 608                      & 132                      & 5992    & 1996  & 832        & 3237        & 665       \\
reproductive system          & -                        & 19                       & 4901    & 656   & 391         & 1307        & 238        \\
respiratory system           & 320                      & 54                       & 2383    & 517   & 148         & 759         & 77         \\
urogenital system            & 721                      & 34                       & 72      & 39    & -           & 109          & -          \\ \hline
\end{tabular}%
}
\end{table*}     

Figures in the database of CRFinder are reconstructed by disease ontology, symptom ontology, and body system ontology. The two statistical figures in Figure 4(A) indicate the number of case reports and figures contained in main root nodes of disease ontology and symptom ontology in CRFinder database. Case reports usually record atypical disease phenotypes. Therefore, we compare the rare disease types in CRFinder database with the Genetic and Rare Diseases Information Center (GARD \footnote{https://rarediseases.info.nih.gov/}), which provides the public with access to current, reliable, and easy-to-understand information about rare or genetic diseases, to demonstrate that CRFinder database has included a certain number of disease types. As indicated in Figure \ref{fig4}(B) and (C), 935 kinds of rare diseases can be found in CRFinder database, which proves the diseases' diversity and educational reference to clinicians.

\subsection{Web interface of CRFinder}
\begin{figure*}[!t]
\centering{\includegraphics[width=0.9\textwidth]{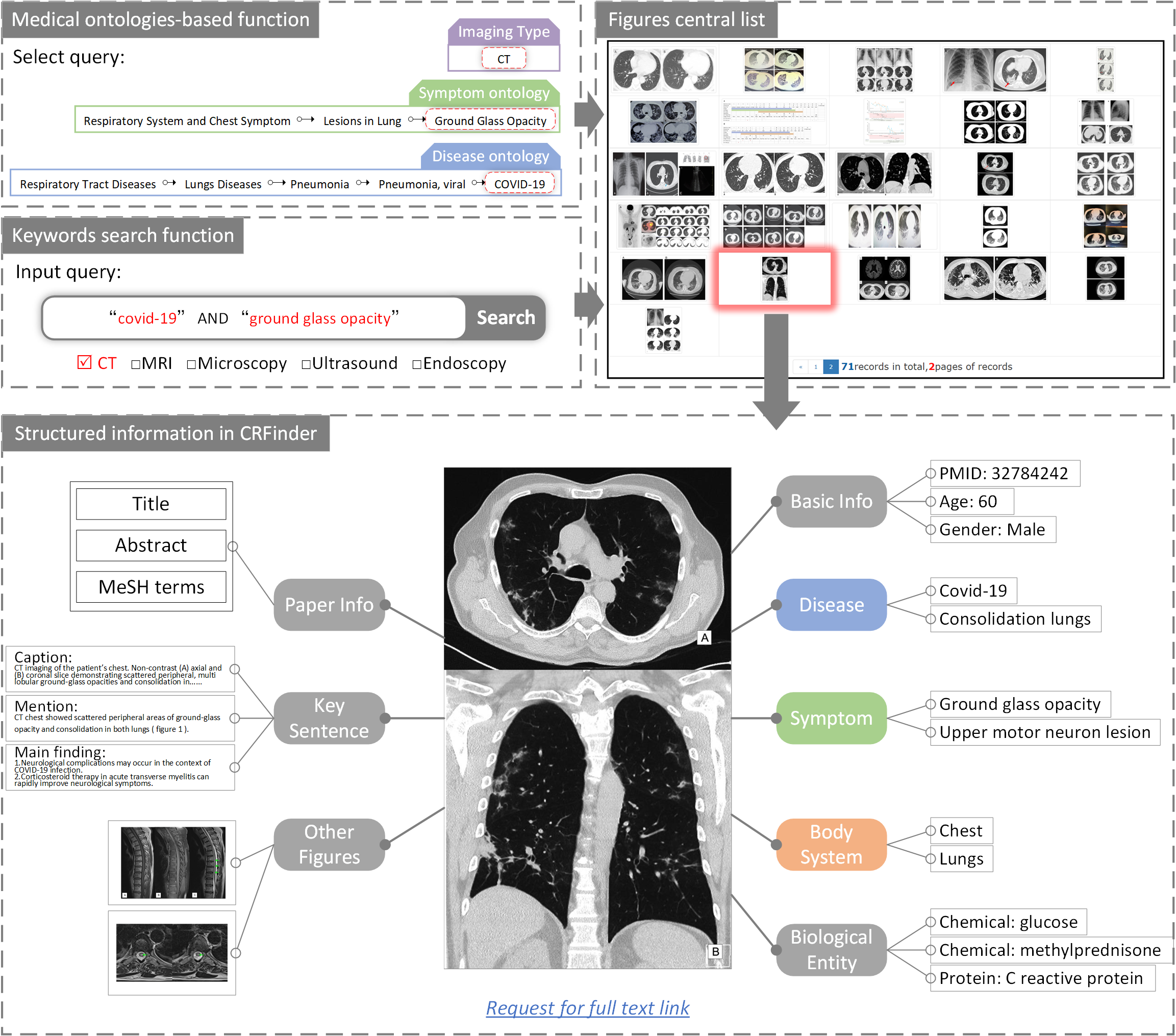}}
\caption{An application case of CRFinder retrieval pipeline. Two query functions: figures browsing through medical ontologies and keyword searching. Figures central list shows the query results. More detailed descriptions are displayed by clicking the figure.}
\label{fig5}
\end{figure*}

Based on CRFinder database, a user-friendly web-based retrieval interface (Figure \ref{fig5}) is designed for searching and analyzing clinical figures with medical entities extracted from case reports. The retrieval interface of CRFinder offers two query modes for clinicians: a medical ontologies-based browse function and a keyword-based search function. When junior clinicians are uncertain about the specific words of diseases, the browse function would provide hints through unfolding nodes layer-by-layer in the medical ontology. On the ‘Browse’ page, users can look through the figure result list based on four filters: disease, symptom, body system, and imaging types. When clinicians would like to request case reports related to specific words or a certain one, the search function on the ‘Search’ page can satisfy clinicians and return results containing query words or the exact PMID of case reports. Users can input a query with multiple keywords, which also supports logic queries such as ‘And’ and ‘Or’. Differences between ‘Browse’ and ‘Search’ results might exist because the browse function will list all figures of child nodes besides selected nodes in medical ontologies while the search function would only display figures that match the keywords in the query. Medical ontology-based retrieval function offers clinicians tips and guidance, which also would support clinicians obtain appropriate references when they are uncertain of some patient conditions. Retrieval results are displayed as figure central lists, each of which has a popover box of brief information (such as age, gender, disease, and symptom). To demonstrate the effectiveness of CRFinder retrieval functions, a use case (PMID:32784242)\cite{b40} is given out in Figure \ref{fig5}.

First, on “Search” page, users can input \emph{“covid-19”} and \emph{“ground glass opacity”} in the query box and choose “CT” imaging type. The search function allows clinicians to search for figures and case reports related to exact keywords, where the specific disease name is known to clinicians. Also, use the other retrieval function in “Browse” page and select keywords directly based on three medical ontologies in the sidebar. Medical ontology-based selection can gradually refine and narrow the retrieval results. This ontology-based browse function could guide and assist clinicians when they are ambiguous about the exact rare disease names. For this application case, processes of selection in disease ontology and symptom ontology are: \emph{“Disease-Respiratory Tract Diseases-Lung Diseases-Pneumonia-Pneumonia, viral-COVID19”} and \emph{“respiratory system and chest symptom-lesions in lung-ground glass opacity”}. Then, all retrieval results under the conditions will be presented in the form of the figure. Secondly, more information will be provided for users by clicking the relevant figure.

On the “Details” page, multimodal structured information will be displayed including medical imaging figures, basic information about patients, clinical entities, biological entities, captions, mentions, and main findings. For more content, full-text links would also lead users to original pages in PubMed where full texts could be found and redirected. Structured information allows clinicians to obtain key information efficiently and decide whether the case report corresponds with their needs. According to our discussion with clinicians on retrieval mode, medical ontology-based selection is consistent with clinicians' preference for retrieving relevant case reports of rare diseases.

\subsection{Result evaluation}
The main workflow of CRFinder contains textual key information extraction (including key sentence and medical entity extraction), medical figures classification, and Case Report retrieval. Captions and mentions are extracted based on XML tag, and medical entity extraction is an unsupervised method based on string cosine similarity. Therefore, in this section, we evaluate the accuracy of main findings extraction, medical figures classification, and case report retrieval results. Moreover, 10 clinicians are invited to use and evaluate the retrieval performance of CRFinder.

\textbf{Evaluation of main findings extraction.} To evaluate the accuracy of main finding sentence extraction in this study, a manual corpus of main findings in case reports \cite{b29}, which contains 500 articles from PubMed, written in English, and published from 1987 to 2017, is chosen as the control group data. From this manual corpus, we filter out 261 case reports which are also collected in CRFinder database. We compare the main finding sentences annotated manually by experts in \cite{b29} and extracted by the step of key information extraction in this study. Specifically, we apply the n-gram algorithm (n=3) to process each main finding sentence and calculate the cosine similarity coefficient between extracted main findings and annotated ones. A threshold (${\alpha}$) of the cosine similarity coefficient is set to calculate the accuracy of main findings extraction. If the cosine similarity coefficient of the sentence is higher than ${\alpha}$, it would be regarded as true positive (TP). If the similarity score is lower than ${\alpha}$, it is the true negative (TN). According to the calculation of (TP + TN)/(total number), we get accuracy values among different cosine similarity coefficient thresholds (Table \ref{table3}).

\begin{table}[!h]
\centering
\caption{Statistics of main findings extraction accuracy among different ${\alpha}$ which is cosine similarity coefficient.\label{table3}}
{%
\begin{tabular}{cccccccc}
\hline
${\alpha}$                       & \textgreater{}0.4 & \textgreater{}0.5 & \textgreater{}0.6 & \textgreater{}0.7 & \textgreater{}0.8 & \textgreater{}0.9 & 1     \\ \hline
Number of main findings & 170               & 165               & 155               & 132               & 108               & 87                & 69    \\
Accuracy                & 0.651             & 0.632             & 0.594             & 0.506             & 0.414             & 0.333             & 0.264 \\ \hline
\end{tabular}%
}
\end{table}

Some main finding sentences express the same meaning but in different word order, which causes low similarity scores by using n-gram algorithm. Figure 6 shows an example of that manually annotated main finding sentence and extracted one that has a low cosine similarity coefficient, but in fact, both describe a spina bifida patient with bladder carcinoma in an auto-augmented bladder. Finally, by manual verification, the threshold of the cosine similarity coefficient is set to 0.5, which has more accurate main findings extraction.

\begin{figure}[!h]
\centering{\includegraphics[width=0.8\columnwidth]{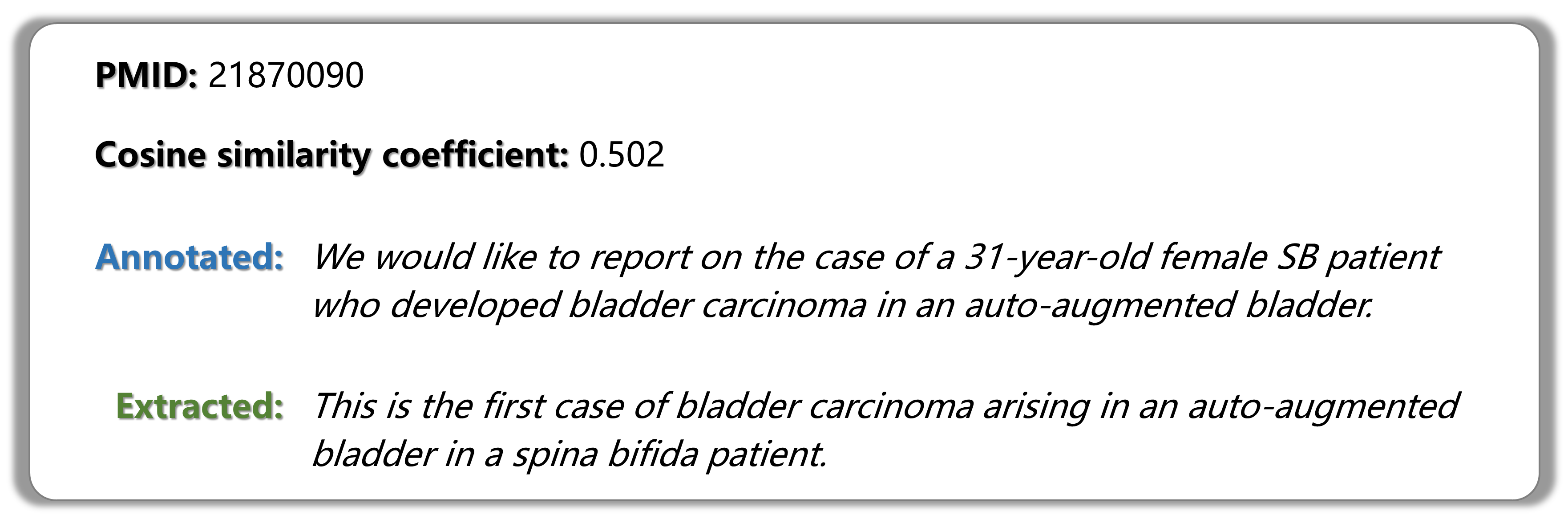}}
\caption{An example of low cosine similarity coefficient between annotation and extraction but the same meaning.}
\label{fig6}
\end{figure}

\begin{figure}[!h]
\centering{\includegraphics[width=0.6\columnwidth]{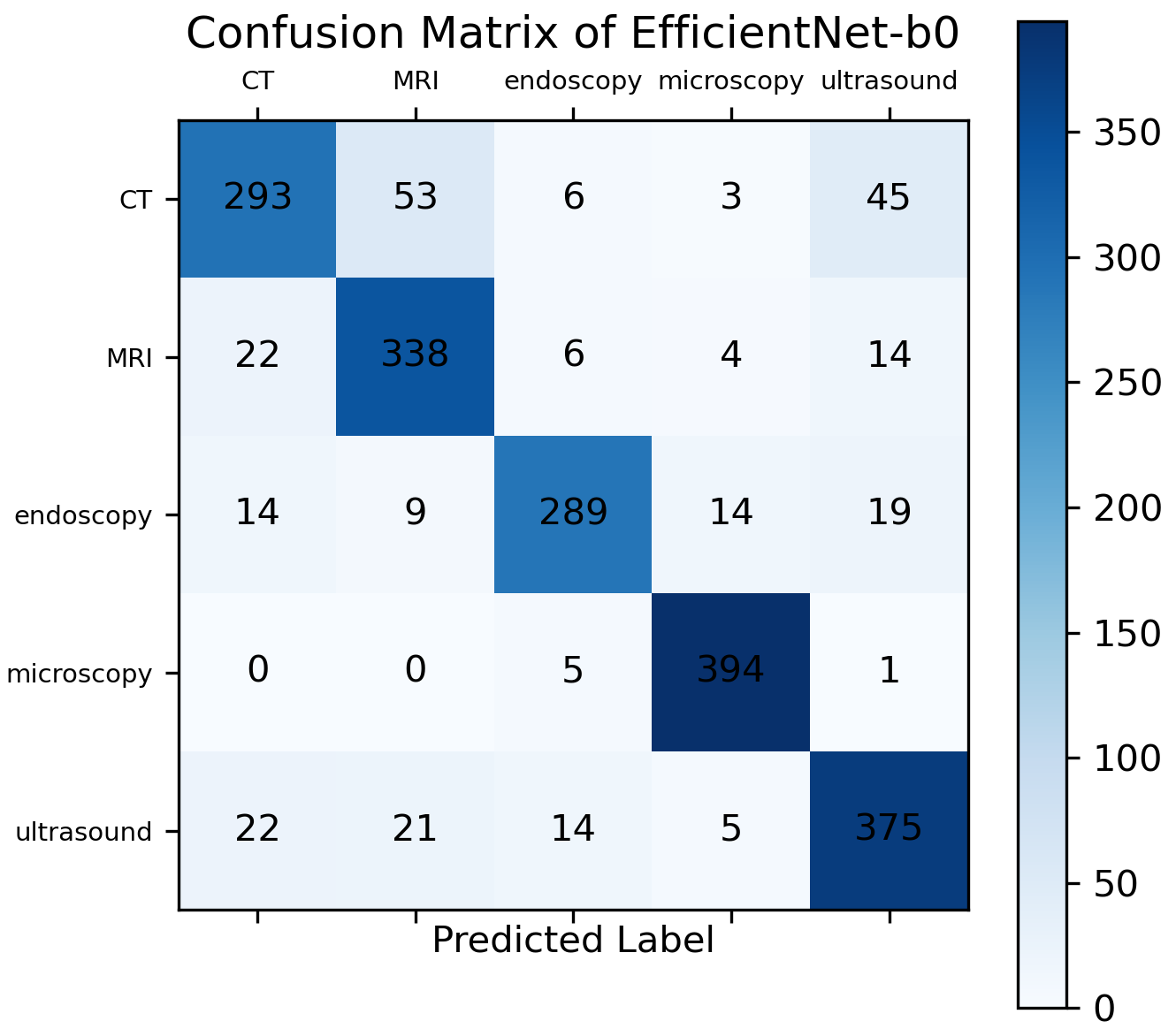}}
\caption{Confusion matrix visualization of EfficientNet-b0.}
\label{fig7}
\end{figure}

\textbf{Evaluation of medical imaging figures classification.} To choose a model with better performance in automatically classifying 5 kinds of medical imaging figures, several popular deep learning models including ResNet, DenseNet, InceptionNet, and EfficientNet are applied in evaluation experiments and the results are shown in Table \ref{table4}. According to the result list, EfficientNet-b0 shows the best accuracy in classifying medical imaging figures which is 0.859 on the test dataset with 1966 figures. Figure \ref{fig7} gives a better visualization of EfficientNet-b0 classification precision on the test dataset.

\begin{table}[!h]
\setlength{\tabcolsep}{14mm}
\centering
\caption{Accuracy results of 7 deep learning models in figures classification. Bold font denotes the highest accuracy.\label{table4}}
\begin{tabular}{cc}
\hline
Model           & Accuracy \\ \hline
ResNet-34       & 0.839    \\
ResNet-50       & 0.832    \\
DenseNet-121    & 0.852    \\
DenseNet-169    & 0.849    \\
Inception-V3    & 0.643    \\
EfficientNet-b0 & \textbf{0.859}    \\
EfficientNet-b7 & 0.854    \\ \hline
\end{tabular}
\end{table}

CRFinder system provides a novel case report retrieval pattern by selecting medical figures in the original literature. We evaluate CRFinder retrieval system on two levels: (1) evaluate retrieval figure results based on the relevant query terms and figure content; (2) evaluate case report retrieval results based on doctors’ actual usage of CRFinder. More details are as follows.

\textbf{Evaluation of CRFinder retrieval figure results.} CRFinder system provides a keywords-based retrieval function to meet clinicians’ personalized searching requirements. To verify the retrieval figure results in CRFinder, we define the evaluation rules as follows: In the retrieval figure results, true positive indicates the contents of figures match the descriptions of query words, and the captions or mentions corresponding to the figures also contain the query words. False negative indicates the descriptions of figures and query words are consistent, and other sentences (e.g. title, abstract, MeSH terms, main findings) contain the query words. 
Likewise, a false positive indicates that the figure contents are not consistent with the query words, but captions or mentions contain the query words. True negative indicates the descriptions of figures and query words do not match, and other sentences that are similar to false negative contain the query words. Though the keyword-based retrieval function may return negative figures, these negative results still have approximate relevance to the queries and provide more expanding information for clinicians. Based above definition, the results ranking strategy is that true positive and false positive results are displayed on the top of the result list, followed by true negative and false negative results.

We invite five clinical doctors with extensive experience and five medical students to use CRFinder system and conduct an evaluation experiment using 100 queries with multiple terms based on the needs of clinical scientific research for case report retrieval. Due to the cost of the manual evaluation, we haven’t constructed more queries. The queries are composed of frequently used terms such as diseases, symptoms, body systems, and clinical imaging procedures. Based on the above definition of positive and negative results, we calculate precision and recall to evaluate the retrieval performance. Part of the query examples and their evaluation results are shown in Table \ref{table5}.

\begin{table*}[]
\centering
\caption{Details of 10 query examples and evaluation of the retrieval figure results.\label{table5}}
{%
\begin{tabular}{ccccc}
\hline
No. & Query Content                                           & Number of figure results & Precision & Recall \\ \hline
1   & (brain) AND (Alzheimer) AND (MRI)                       & 25                       & 0.90      & 0.39   \\
2   & (lung) AND (cough) AND (CT)                             & 440                      & 0.89      & 0.22   \\
3   & (colon cancer) AND (Endoscopy)                          & 16                       & 1.00      & 0.55   \\
4   & (Breast tumor) AND (Microscopy)                         & 29                       & 0.94      & 0.61   \\
5   & (polydactyly) AND (skin hyperpigmentation) AND (anemia) & 1                        & 0.00      & 0.00   \\
6   & ((cough) OR (fever)) AND (Covid-19) AND (CT)            & 451                      & 0.93      & 0.16   \\
7   & (Autoimmune Encephalitis) AND (MRI)                     & 18                       & 0.80      & 0.27   \\
8   & (Parkinson) AND (young)                                 & 21                       & 1.00      & 0.11   \\
9   & (Retinoblastoma) AND (eye)                              & 30                       & 1.00      & 0.29   \\
10  & (Thomsen) OR (Becker) OR (myotonia)                     & 56                       & 0.70      & 0.21   \\ \hline
\multicolumn{2}{c}{AVG.}                                      & 108.7                    & 0.82      & 0.28   \\ \hline
\end{tabular}%
}
\end{table*}

The results indicate that CRFinder retrieval system performs a good performance on figures retrieval with precisions and recalls of average 0.816 and 0.281. As for the No.5 query, the three words are typical symptoms of Fanconi anemia disease. Though the No.5 query has 0 precision and recall for retrieval evaluation, the figure of this result is highly relevant to the query. From numerical evaluation, CRFinder retrieval system does not show better precision or recall, and the number of true positive and false negative results is a high proportion of overall results, which means most of the figures are related to query and could effectively improve doctors’ actual retrieval on case reports.

\textbf{Evaluation of CRFinder case report retrieval results.} To assess the case report retrieval accuracy of CRFinder, we conduct an additional evaluation experiment inviting five senior doctors and five postgraduate medical students. They are asked to input query words and select case reports from the results. By clicking figures and viewing the Details page, if the Case Report is the exact or approximate they want, the case report would be regarded as a true positive result. Finally, we count the number of doctors and medical students’ selections in the retrieval results, which are shown in Table \ref{table6}.

\begin{table}[]
\centering
\caption{Results of recording doctors' and medical students' queries and the number of case reports satisfying doctors in the retrieval results. ${Num^{Q}}$ is the number of queries per user. ${Num^{Fig}}$ is the number of figures returned by CRFinder. ${Num^{CR}}$ is the number of case reports that users consider relevant to their queries.\label{table6}}
\begin{tabular}{cccc}
\hline
Users               & ${Num^{Q}}$  & ${Num^{Fig}}$   & ${Num^{CR}}$   \\ \hline
Doctor \#1          & 5  & 75  & 12  \\
Doctor \#2          & 8  & 120 & 32  \\
Doctor \#3          & 10 & 150 & 21  \\
Doctor \#4          & 9  & 135 & 20  \\
Doctor \#5          & 6  & 90  & 9   \\
Medical Student \#1 & 15 & 225 & 81  \\
Medical Student \#2 & 10 & 150 & 65  \\
Medical Student \#3 & 14 & 210 & 103 \\
Medical Student \#4 & 12 & 180 & 77  \\
Medical Student \#5 & 11 & 165 & 71  \\ \hline
\end{tabular}
\end{table}

From the results, we notice that doctors and medical students could find relevant figures and corresponding case reports according to their queries. After communication with doctors and medical students, doctors seem to get less even non-relevant figures than medical students. The main reason is that senior doctors prefer case reports with specific diseases or treatments and have clearer retrieval direction. For medical students, the query words are fuzzier and their requirements focus on retrieving a series of case reports about some cases with rare clinical symptoms or treatments for complex diseases. Compared to the textual results list of other literature retrieval platforms, browsing figure central results in CRFinder could efficiently assist medical students or any other junior clinicians in case report retrieval. Medical figures convey a more direct and irreplaceable message to doctors than texts.

\begin{table}[]
\setlength{\tabcolsep}{5mm}
\centering
\caption{Comparison of PubMed, YIF, SIBiLS, Open-i, and CRFinder.\label{table7}}
{%
\begin{tabular}{cccccc}
\hline
Tool Name & PubMed & YIF & SIBiLS & Open-i & CRFinder \\ \hline
Data Source & PubMed & PMC & PMC & PMC & PMC \\ 
Figure Browse Function& No & Yes & No & Yes & Yes \\   
Medical Imaging Type Filters & None & None & None & 8 & 5 \\ 
Medical Ontology Guidance & No & No & No & No & Yes \\ 
Include Case Report & Yes & No & No & Yes & Yes \\ 
\begin{tabular}[c]{@{}c@{}}Personalized Retrieval on \\ Case Report\end{tabular} & No & No & No & No & Yes \\
Content Scope& \multicolumn{1}{l}{\begin{tabular}[t]{@{}l@{}}Title\\ Abstract\end{tabular}} & \multicolumn{1}{l}{\begin{tabular}[t]{@{}l@{}}Title\\ Abstract\\ Caption\end{tabular}} & \multicolumn{1}{l}{\begin{tabular}[t]{@{}l@{}}Title\\ Abstract\\ Caption\\ MeSH\end{tabular}} & \multicolumn{1}{l}{\begin{tabular}[t]{@{}l@{}}Title\\ Abstract\\ Caption\\ Mention\end{tabular}} & \multicolumn{1}{l}{\begin{tabular}[t]{@{}l@{}}Title\\ Abstract\\ Caption\\ Mention\\ MeSH\end{tabular}} \\
 \hline
\end{tabular}%
}
\end{table}

\section{Discussion}
We perform a comparison of CRFinder, PubMed, Yale Image Finder (YIF) \cite{b41}, Swiss Institute of Bioinformatics Literature Services (SIBiLS) \cite{b42}, and Open-i \cite{b26} among 7 attributes (Table \ref{table7}). Each attribute is considered when doctors expect a system to bring them during case report retrieval.

Collections of most existing medical literature retrieval tools are biomedical research papers and some of them may also include case reports, which are designed for more accurate literature retrieval results, but the demands and preferences of clinicians when searching for case reports have not been satisfied perfectly. PubMed is recognized as the most powerful platform for searching medical literature and it has included plenty of literature of different types. Comparatively, CRFinder has a novel medical ontology-based retrieval function and a figure-based results displayed format which provides a much more efficient way of case report retrieval than PubMed. Due to the limitation of literature copyright, CRFinder only includes case reports with full paper free open access. This is a deficiency of CRFinder system and does not contain as much literature as PubMed. SIBiLS allows full customization search in extra full texts and only provides textual result lists. By communicating with clinicians, medical figures in case reports always indicate direct expression about patient conditions which cannot be described by textual sentences. YIF and Open-i allow users to search literature through browsing figures. Additionally, Open-i provides medical imaging type filters to screen specific imaging procedure figures. However, novice or junior doctors may be confused when meeting patients with rare diseases or some unusual symptoms and need to seek references through case reports or relevant literature. Under this statement, compared with Open-i query figures through inputting textual words or sentences, medical ontologies including disease, symptom, and body system are constructed as guidance in CRFinder information system. Clinicians could be uncertain of the specific disease names facing patients with unusual phenotypes and need to review case reports by matching similar clinical manifestations. Medical ontology guidance in CRFinder could mitigate this deficiency by hinting clinicians through hierarchical clinical entities. With non-commercial use license of PMC open access subset, CRFinder has collected 58601 medical figures from case reports published from 2000 to 2021 and extracted structured summary (such as entities of disease, symptom, and body system), which is more comprehensive than Open-i. Moreover, CRFinder is customized for clinicians to retrieve case reports rather than general biomedical literature more efficiently by reviewing structured multimodal information. The retrieval function in CRFinder still needs to be improved by adding semantic text query, collecting more diverse and accurate structured information, fasting retrieval efficiency, and so on. We would continue to optimize and extend CRFinder database.

\section{Conclusion}
According to our research, CRFinder is the first comprehensive database focusing on case reports and assisting junior doctors, which has integrated multimodal summaries from case reports including (i) medical imaging figures (such as CT, MRI, ultrasonography, microscopy, and endoscopy); (ii) clinical disease, symptom, and body system entities; (iii) biological gene, chemical, protein entities; (iv) key sentences. Meanwhile, a user-friendly web interface is provided for clinicians to achieve an efficient and convenient retrieval, visualization, and analysis of case reports by utilizing a medical ontology-based browsing function as guidance, which would enhance clinicians’ identification of cases with atypical disease phenotypes. The current version of CRFinder contains 52949 case reports which are freely downloaded in the open-access subset of PubMed Central. There will be more supplements of case reports and more accurate retrieval results for CRFinder and web platforms in the future. Taken together, this study is not only beneficial for clinicians in searching case reports but also provides an opportunity for extracting multimodal case information from medical literature.

\bibliographystyle{unsrt}  
\bibliography{references}

\end{document}